\documentclass[11pt]{article}
\usepackage[preprint]{acl}
\usepackage{times}
\usepackage{latexsym}
\usepackage[T1]{fontenc}
\usepackage[utf8]{inputenc}
\usepackage{microtype}
\usepackage{inconsolata}
\usepackage{graphicx}
\usepackage{booktabs}
\usepackage{multirow}
\usepackage{xcolor}
\usepackage{colortbl}
\usepackage{amsmath,amssymb}
\usepackage{pifont}
\usepackage{tabularx}
\usepackage{placeins}
\usepackage{algorithm}
\usepackage[noend]{algpseudocode}
\usepackage{tikz}
\usepackage{pgfplots}
\pgfplotsset{compat=1.18}
\usetikzlibrary{arrows.meta,positioning,calc}
\definecolor{oursrow}{HTML}{E2F7FE}
\newcolumntype{W}{!{\color{white}\vrule width 0.84pt}}
\definecolor{projected}{HTML}{0D63FF}
\definecolor{gaindn}{HTML}{808080}
\definecolor{gainup}{HTML}{0D63FF}
\definecolor{policyblue}{HTML}{087AA3}
\definecolor{languageblue}{HTML}{DDEAF7}
\definecolor{ink}{HTML}{253449}
\newcommand{\method}{MedDEx}
\newcommand{\rx}{$^{\star}$}
\algrenewcommand\algorithmicrequire{\textbf{Input:}}
\algrenewcommand\algorithmicensure{\textbf{Output:}}

\title{More Reasoning Is Not Always Better: Executing\\Clinical Guidelines with LLMs and Rules}
\author{
  Xiaoqian Lu$^{1}$ \quad
  Guangfu Guo$^{2}$ \quad
  Long Cheng$^{2}$ \quad
  Kunpeng Liu$^{2}$ \\[5pt]
  \normalfont
  $^{1}$University of Ottawa \quad
  $^{2}$Clemson University
}

\begin{document}
\maketitle
\begin{abstract}
Large language models (LLMs) have shown considerable promise in clinical decision support. However,
existing approaches often rely on multi-step reasoning or multi-agent collaboration to execute clinical
guidelines, introducing substantial computational overhead without reliably preventing decision errors.
To address these limitations, we propose a framework for \textbf{Med}ical \textbf{D}ecision
\textbf{Ex}ecution (\method{}) that integrates structured patient states with deterministic rule
execution to support guideline-based treatment planning. \method{}
separates clinical record interpretation from guideline execution, leveraging LLMs to extract patient
information and generate treatment plans while assigning guideline-defined decision logic to a rule
executor. Experiments across multiple clinical guideline benchmarks demonstrate that \method{} improves
decision accuracy while reducing inference costs. Further analyses highlight the critical roles of
structured patient representations and deterministic rule execution, suggesting that additional LLM
reasoning does not necessarily translate into better guideline adherence.
\end{abstract}

\section{Introduction}

Large language models (LLMs) have demonstrated immense potential in clinical decision support, capable of extracting key information from complex clinical records and generating personalized treatment recommendations~\citep{thirunavukarasu2023llm,singhal2023medpalm,singhal2025expert,benary2023oncology}. However, guideline-based treatment planning requires not only an understanding of patient information but also the accurate execution of decision logic prescribed by clinical guidelines. For instance, in oncology, the selection of a treatment regimen often depends on factors such as disease stage, molecular biomarkers, and prior treatment history, with corresponding decision pathways explicitly defined in clinical guidelines~\citep{riely2024nccn,chen2025csco}. Despite their strong language understanding and reasoning capabilities, LLMs may still select incorrect treatment pathways during guideline execution, even when relevant clinical information is correctly identified~\citep{chen2023chatbots,hager2024limitations,hao2026lungcure}.

\begin{figure}[t]
\centering
\includegraphics[width=\columnwidth]{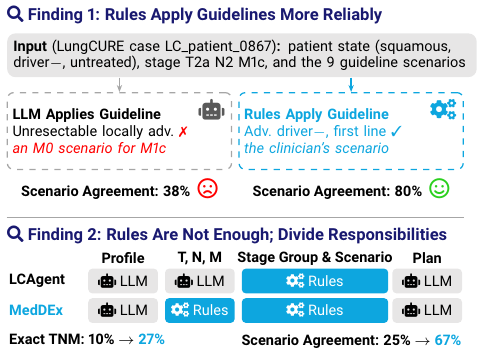}
\caption{\textbf{The importance of guideline execution and the division of responsibilities.} Top: Deterministic rule execution is more reliable than LLM-based guideline execution given matched inputs. Bottom: \method{} improves decision performance by refining routing inputs and allocating responsibilities for TNM computation.}
\label{fig:motivation}
\end{figure}

To enhance the reliability of clinical decision-making, researchers have proposed various LLM-based reasoning enhancement methods. Retrieval-augmented generation (RAG) supports decision-making by incorporating external medical knowledge~\citep{lewis2020rag,xiong2024medrag}; self-reflection improves reasoning outcomes by iteratively refining model outputs~\citep{shinn2023reflexion}; and multi-agent collaboration improves clinical recommendations through discussion and feedback among different agents~\citep{tang2023medagents,kim2024mdagents,du2024debate}. Additionally, some studies have attempted to enhance the stability of the reasoning process through structured intermediate representations~\citep{li2025drs,xiong2025mapping}. Although these methods have shown effectiveness across various tasks, most still rely on LLMs themselves to execute the decision logic defined in clinical guidelines.

However, existing methods face three key challenges. First, additional reasoning does not necessarily improve guideline execution. Even with identical patient states and disease stages, LLMs may select decision paths inconsistent with guideline rules (Figure~\ref{fig:motivation}, top). Second, multi-round reflection and multi-agent collaboration incur substantial computational overhead by repeatedly deriving decision logic already defined in clinical guidelines. Third, introducing rule executors alone is insufficient, as their reliability depends on accurate input states. Existing methods often rely on LLM-generated free-text patient profiles or predicted disease stages, allowing intermediate errors to propagate into final decisions (Figure~\ref{fig:motivation}, bottom). Therefore, accurately representing patient information while ensuring reliable and efficient guideline execution remains a critical challenge.

\begin{figure*}[t]
\centering
\includegraphics[width=0.95\textwidth]{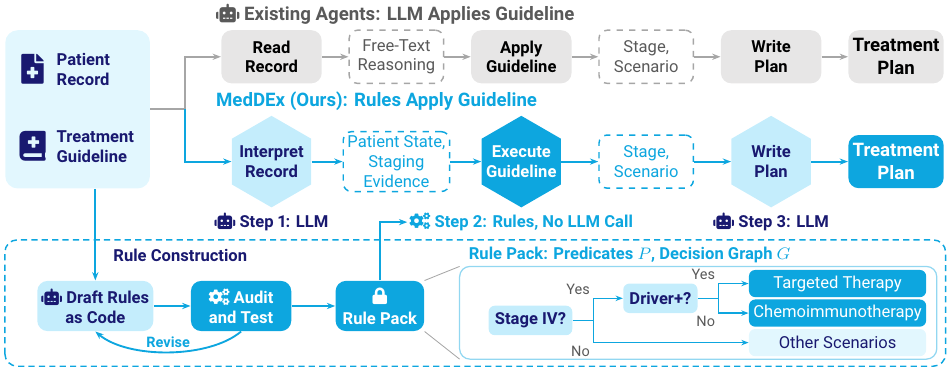}
\caption{\textbf{Overview of \method{}.} Top: LLM-executed agents assign scenario selection to the model. Middle: \method{} assigns record interpretation and plan generation to the LLM, and TNM computation and scenario routing to rules. Bottom (dashed): coding assistants draft rule packs, audits and tests guide revision, and the authors review the code before evaluation. The graph is simplified.}
\label{fig:method}
\end{figure*}

To address these challenges, we propose \textbf{\method{}}, a guideline-based treatment planning framework that integrates structured patient states with deterministic rule execution. We first develop a rule executor by translating guideline-defined decision conditions and pathways into executable rules. As shown in Figure~\ref{fig:method}, the inference process consists of three stages. First, an LLM extracts key information from clinical records and constructs structured patient states. Next, the rule executor computes derived conditions and selects appropriate treatment pathways. Finally, the LLM generates treatment recommendations based on the selected guidelines. By clearly dividing responsibilities between LLMs and rule executors, \method{} enables reliable and efficient guideline execution without additional model calls for decision-making.

We conducted extensive evaluations of \method{} on three clinical guideline benchmarks: MedGUIDE~\citep{li2025medguide}, MEGA-CDP~\citep{chen2026megacdp}, and the bilingual Chinese--English LungCURE benchmark~\citep{hao2026lungcure}. Experimental results demonstrate that \method{} improves guideline decision agreement and staging accuracy while reducing inference costs. Further ablation studies validate the importance of structured patient states and deterministic rule execution, indicating that appropriately dividing responsibilities between LLMs and rule executors enables more accurate and efficient guideline-based decision-making, whereas additional LLM reasoning does not always yield better results.

The main contributions of this paper are as follows:
\begin{enumerate}
\itemsep0pt
    \item We identify guideline execution as a distinct source of error and show that additional LLM reasoning does not necessarily eliminate these errors.
    \item We propose \method{}, which integrates structured patient states with deterministic rule execution by clearly dividing responsibilities between LLMs and rule executors.
    \item We demonstrate the effectiveness and efficiency of \method{} across multiple clinical guideline benchmarks, supported by systematic ablation studies.
\end{enumerate}

\section{Related Work}

\subsection{Medical LLM Agents}

The clinical knowledge capabilities of large language models (LLMs) have driven the development of medical agents~\citep{singhal2023medpalm,thirunavukarasu2023llm}. MedAgents performs clinical reasoning through collaboration among specialist agents~\citep{tang2023medagents}, while MDAgents dynamically adjusts the agent team size based on case complexity~\citep{kim2024mdagents}. AgentClinic and MedAgentBench evaluate agent capabilities in interactive clinical environments and electronic health record (EHR) tasks, respectively~\citep{schmidgall2024agentclinic,jiang2025medagentbench}. For lung cancer treatment planning, LungCURE introduces LCAgent, which makes decisions through evidence extraction, patient profile construction, rule-based scenario routing, and guideline-conditioned treatment plan generation~\citep{hao2026lungcure}. These agents differ in how they organize the model, but all of them leave the guideline decision to it.

\subsection{Enhancing LLM Reasoning}

Existing methods enhance model performance by adding reasoning steps~\citep{wei2022cot,wang2023selfconsistency,guo2026deltaflow}, combining reasoning with tool use~\citep{yao2023react}, and employing self-reflection~\citep{shinn2023reflexion,guo2026beyond}. Retrieval-augmented generation (RAG) introduces external knowledge~\citep{lewis2020rag,xiong2024medrag}; MA-RAG conducts multi-round retrieval based on disagreements among candidate answers~\citep{wu2026marag}, and DermAgent completes tasks by integrating a planner, executor, and critic~\citep{liu2026dermagent}. A separate line constrains what the model emits, reformulating questions or reasoning traces into structured output so that later steps rest on something stable~\citep{li2025drs,xiong2025mapping,guo2025structured}. While multi-agent debate can improve factual accuracy, agreement on final answers does not imply consistency in the underlying reasoning processes~\citep{du2024debate,wang2026consistency}. Although these methods enhance performance by incorporating additional information, reasoning steps, or multi-agent collaboration, guideline-defined decisions are still primarily made by LLMs.

\subsection{Executable Clinical Guidelines}

Research on translating clinical guidelines into executable logic has a long history. Early frameworks, such as the Arden Syntax and PROforma, executed clinical decision rules based on structured data~\citep{hripcsak1994arden,sutton2003proforma,peleg2003comparing}. With the advent of LLMs, one line of work guides models to evaluate conditions step by step along guideline decision trees~\citep{li2023meddm,deng2026cpgprompt} or internalizes guideline decision logic directly into the models~\citep{shen2026medguidex}; another line of work utilizes LLM-generated inputs to invoke clinical calculators~\citep{jin2025agentmd}, emergency decision rules~\citep{xiang2025cdragent}, and guideline conflict solvers~\citep{xie2026neurosymbolic}. Additionally, Guideline2Graph converts clinical guidelines into decision graphs~\citep{kilic2026guideline2graph}, while PAL and Logic-LM combine LLMs with program interpreters or symbolic solvers to perform reasoning~\citep{gao2023pal,pan2023logiclm}. In contrast to these studies, this paper compares LLM-based guideline execution with deterministic rule execution under matched inputs, and investigates how responsibilities should be divided between the two.

\section{Methodology}
\label{sec:method}

MedDEx decomposes guideline-based treatment planning into three stages: clinical record interpretation by a large language model (LLM), deterministic guideline execution by a rule executor, and guideline-conditioned answer generation by the LLM. As illustrated in Figure~\ref{fig:method}, this framework establishes a clear division of responsibilities between language understanding and rule execution.

\subsection{Decision Process}
\label{sec:process}

Given a clinical record $x$ and a treatment guideline, the objective is to generate a guideline-consistent answer $y$. We represent each guideline as a rule pack:
\begin{equation}
\mathcal{B}=(P,\mathcal{F},G)
\end{equation}
where $P=\{p_1,\ldots,p_m\}$ is the set of guideline decision conditions, $\mathcal{F}$ contains rules for computing derived conditions, and $G=(V,E)$ is a guideline decision graph whose branches evaluate conditions to determine decision paths. Each decision is associated with corresponding guideline content. For guidelines without derived conditions, $\mathcal{F}$ is empty.

Existing LLM-executed agents typically rely on the model to interpret clinical records, select guideline pathways, and generate answers directly:
\begin{equation}
y=\mathcal{M}(x,G)
\end{equation}
In contrast, MedDEx separates this process into clinical record interpretation, deterministic guideline execution, and guideline-conditioned generation:
\begin{equation}
\begin{aligned}
(\hat{s},e)&=\mathcal{M}(\textsc{Read},x,P),\\
s&=\hat{s}\cup\mathcal{F}(e),\\
O&=R_{\mathcal{B}}(s),\\
y&=\mathcal{M}(\textsc{Write},x,s,G_O).
\end{aligned}
\end{equation}
Here, $\mathcal{M}$ denotes the LLM, $\hat{s}$ represents the extracted condition values, $e$ denotes the raw clinical evidence, and $s$ is the patient state augmented with derived conditions. $R_{\mathcal{B}}$ is the rule executor, $O$ is the set of guideline decisions returned by execution and $G_O$ denotes the corresponding guideline content. Only the interpretation and generation stages require LLM calls, while guideline decisions are executed by deterministic rules.

\subsection{Patient Interpretation Module}
\label{sec:interpreter}

This module uses the LLM to map unstructured clinical records into structured condition values required by the guideline. For each directly extractable condition $p_i$, the extracted value satisfies:
\begin{equation}
\hat{s}_i\in\mathcal{V}_i,
\quad i=1,\ldots,m
\end{equation}
where $\mathcal{V}_i$ denotes the predefined value set for condition $p_i$, following the design of each benchmark. For conditions that must be derived from clinical evidence, the LLM extracts only the raw evidence $e$, leaving the corresponding computation to the rule executor rather than directly producing guideline decisions.

This design allows the LLM to focus on clinical record interpretation without directly performing decisions explicitly defined by the guideline.

\subsection{Rule Executor}
\label{sec:executor}
\label{sec:construction}

\paragraph{Rule Construction.} We develop a rule pack $\mathcal{B}$ for each clinical guideline, comprising decision conditions, derived-condition rules, and a decision graph. Specifically, GPT-6-Astra and Claude Opus 5.5 assist in generating executable rule code, which is subsequently checked and revised through structural audits, guideline-based example tests, and manual author review. All rule packs are frozen before evaluation. Further construction details are provided in Appendix~\ref{app:external}.

\paragraph{Deterministic Execution.} Given the patient state $s$, the rule executor first computes derived conditions and then evaluates branch conditions along the decision graph. Starting from the entry node $v_0$, the execution process is formalized as:
\begin{equation}
v_{t+1}=\delta_G(v_t,s),
\qquad
\pi=(v_0,v_1,\ldots,v_T)
\end{equation}
where $\delta_G$ denotes the guideline-defined transition function, and $\pi$ represents the execution path.

Upon reaching a terminal node $v_T$, the executor returns the set of guideline decisions permitted at that node:
\begin{equation}
O=\lambda(v_T)\subseteq\mathcal{O}
\end{equation}
where $\mathcal{O}$ denotes the guideline decision space. When the guideline specifies a unique decision, $|O|=1$; when multiple options remain valid, the generation module selects among the permitted candidates. Since the execution process involves no LLM calls, identical inputs always produce the same decision path, making intermediate decisions traceable and inspectable.

\subsection{Guideline-Conditioned Generation Module}
\label{sec:planner}

After rule execution, the LLM generates the final answer based on the patient state, execution results, and corresponding guideline content:
\begin{equation}
y=\mathcal{M}(\textsc{Write},x,s,G_O)
\end{equation}
Unlike methods that allow the LLM to select pathways directly from the complete guideline, this module generates answers only within the decision space $O$ determined by the rule executor, without re-executing guideline decisions.

\begin{table*}[t]
\centering\small
\setlength{\tabcolsep}{4pt}
\begin{tabular*}{\textwidth}{@{\extracolsep{\fill}}l WrWrWr WrWrWr@{}}
\toprule
& \multicolumn{3}{c}{\textbf{Qwen3-8B}} & \multicolumn{3}{c}{\textbf{Llama-3.1-8B-Instruct}} \\
\cmidrule(lr){2-4}\cmidrule(lr){5-7}
& MedGUIDE & \multicolumn{2}{c}{MEGA-CDP} & MedGUIDE & \multicolumn{2}{c}{MEGA-CDP} \\
\cmidrule(lr){2-2}\cmidrule(lr){3-4}\cmidrule(lr){5-5}\cmidrule(lr){6-7}
\textbf{Method} & Acc. & Concl. & Path OT & Acc. & Concl. & Path OT \\
\midrule
Single LLM & 68.77 & 83.17 & \underline{77.60} & 66.65 & 74.31 & 71.70 \\
Guideline-RAG & 51.68 & 76.00 & 76.95 & 54.78 & 67.41 & \underline{72.89} \\
MedAgents~\citep{tang2023medagents} & \underline{69.36} & 77.15 & 73.72 & 56.09 & 67.34 & 64.92 \\
MDAgents~\citep{kim2024mdagents} & 66.03 & 78.64 & 72.84 & 50.29 & 65.38 & 69.00 \\
MA-RAG & 69.27 & \underline{83.71} & 75.84 & \underline{68.94} & \underline{74.97} & 71.46 \\
DermAgent~\citep{liu2026dermagent} & 52.43 & 72.70 & 71.11 & 58.89 & 64.33 & 68.20 \\
\rowcolor{oursrow} \method{}\rx{} (ours) & \textbf{69.70} & \textbf{96.29} & \textbf{80.74} & \textbf{69.25} & \textbf{92.24} & \textbf{79.12} \\
\bottomrule
\end{tabular*}
\caption{\textbf{LLM execution vs.\ rule execution of the same guideline.}
MedGUIDE reports option accuracy (\%); MEGA-CDP reports conclusion agreement (\%) and path
similarity (\%).
Best in bold, second underlined; the shaded row is ours.
$^{\star}$: Rules execute the guideline.}
\label{tab:external}
\end{table*}

\begin{table*}[t]
\centering\small
\setlength{\tabcolsep}{3.2pt}
\resizebox{\textwidth}{!}{%
\begin{tabular}{l WcWc WcWc WcWc WcWc WcWc WcWc}
\toprule
& \multicolumn{4}{c}{\textbf{TNM Staging}} & \multicolumn{4}{c}{\textbf{Treatment Recommendation}} & \multicolumn{4}{c}{\textbf{End-to-End Decision Support}} \\
\cmidrule(lr){2-5}\cmidrule(lr){6-9}\cmidrule(lr){10-13}
& \multicolumn{2}{c}{Acc(\%)$\uparrow$} & \multicolumn{2}{c}{RQ$\uparrow$} & \multicolumn{2}{c}{Precision(\%)$\uparrow$} & \multicolumn{2}{c}{F1$\uparrow$} & \multicolumn{2}{c}{Precision(\%)$\uparrow$} & \multicolumn{2}{c}{F1$\uparrow$} \\
\cmidrule(lr){2-3}\cmidrule(lr){4-5}\cmidrule(lr){6-7}\cmidrule(lr){8-9}\cmidrule(lr){10-11}\cmidrule(lr){12-13}
\textbf{Method} & ZH & EN & ZH & EN & ZH & EN & ZH & EN & ZH & EN & ZH & EN \\
\midrule
\multicolumn{13}{l}{\textit{Base model: Qwen2.5-7B-Instruct}} \\
Direct prompting & 0.10 & 2.20 & 33.24 & 36.56 & 38.19 & 57.05 & 41.38 & \underline{52.86} & 40.98 & 55.51 & 39.80 & 45.48 \\
Guideline-RAG & 2.70 & 2.60 & 39.77 & 38.91 & 36.77 & 48.32 & 43.88 & 49.90 & 40.03 & 48.64 & 42.60 & 46.28 \\
MedAgents~\citep{tang2023medagents} & 5.40 & 6.30 & 46.04 & \underline{51.28} & 44.09 & 55.99 & \underline{51.38} & 51.42 & 40.58 & 45.76 & 47.70 & \underline{49.16} \\
MDAgents~\citep{kim2024mdagents} & 2.70 & 4.30 & 41.69 & 50.34 & 44.52 & 53.46 & 51.12 & 50.34 & 44.43 & 50.28 & 47.18 & 46.62 \\
MA-RAG~\citep{wu2026marag} & 3.20 & 3.60 & 41.34 & 39.13 & 45.68 & \underline{59.23} & 43.64 & 48.22 & 47.16 & \underline{58.78} & 43.52 & 42.90 \\
DermAgent~\citep{liu2026dermagent} & 9.00 & 7.30 & 43.51 & 47.91 & \underline{48.83} & 51.55 & 47.32 & 47.68 & 49.22 & 51.93 & 43.80 & 45.66 \\
LCAgent\rx~\citep{hao2026lungcure} & \underline{12.40} & \underline{10.20} & \underline{57.80} & 49.55 & 43.43 & 39.52 & 50.18 & 45.12 & \underline{49.49} & 42.29 & \underline{48.88} & 39.48 \\
\addlinespace[1pt]
\rowcolor{oursrow} \method{}\rx{} (ours) & \textbf{24.40} & \textbf{26.90} & \textbf{70.74} & \textbf{70.78} & \textbf{71.97} & \textbf{73.90} & \textbf{71.04} & \textbf{65.50} & \textbf{67.93} & \textbf{70.58} & \textbf{60.48} & \textbf{57.62} \\
\midrule
\multicolumn{13}{l}{\textit{Base model: Qwen3-8B}} \\
Direct prompting & 2.30 & 3.40 & 41.89 & 44.02 & \underline{47.49} & 47.86 & 54.46 & 51.70 & \underline{47.46} & 53.36 & 53.44 & \underline{49.56} \\
Guideline-RAG & 6.20 & 3.40 & 47.67 & 48.36 & 36.15 & 48.67 & 46.18 & 43.98 & 33.45 & 49.82 & 40.86 & 36.80 \\
MedAgents~\citep{tang2023medagents} & 11.00 & 10.90 & 52.28 & 50.96 & 46.55 & 49.43 & 50.44 & 49.36 & 44.00 & 51.17 & 46.18 & 47.66 \\
MDAgents~\citep{kim2024mdagents} & 9.60 & 10.50 & 53.01 & 53.32 & 45.07 & 50.89 & 47.68 & 49.14 & 45.77 & 55.87 & 46.62 & 44.04 \\
MA-RAG~\citep{wu2026marag} & 7.50 & 4.00 & 50.19 & 44.07 & 34.96 & 53.59 & 45.88 & 44.26 & 31.75 & 55.27 & 41.36 & 40.40 \\
DermAgent~\citep{liu2026dermagent} & 12.80 & 10.00 & 51.80 & 49.92 & 44.36 & 49.41 & 46.00 & 46.10 & 43.32 & 50.81 & 43.30 & 44.14 \\
LCAgent\rx~\citep{hao2026lungcure} & \underline{22.10} & \underline{18.30} & \underline{70.19} & \underline{63.56} & 46.45 & \underline{65.92} & \textbf{65.62} & \underline{58.42} & 45.14 & \underline{57.07} & \textbf{57.38} & 49.18 \\
\addlinespace[1pt]
\rowcolor{oursrow} \method{}\rx{} (ours) & \textbf{30.30} & \textbf{31.90} & \textbf{72.54} & \textbf{71.53} & \textbf{68.62} & \textbf{70.61} & \underline{60.94} & \textbf{62.50} & \textbf{63.52} & \textbf{65.18} & \underline{53.90} & \textbf{55.76} \\
\midrule
\multicolumn{13}{l}{\textit{Base model: Llama-3.1-8B-Instruct}} \\
Direct prompting & 0.30 & 1.70 & 30.75 & 36.87 & 18.18 & 10.86 & 38.08 & 2.76 & 9.07 & 44.67 & 28.06 & 43.06 \\
Guideline-RAG & 2.80 & 1.50 & 39.92 & 42.17 & 23.73 & 42.72 & 36.80 & 38.38 & 24.66 & 38.72 & 34.84 & 36.28 \\
MedAgents~\citep{tang2023medagents} & 6.30 & \underline{7.40} & \underline{50.88} & \underline{52.37} & \underline{38.27} & 54.83 & 37.92 & 44.18 & \underline{34.43} & 53.89 & \underline{37.84} & \underline{43.86} \\
MDAgents~\citep{kim2024mdagents} & 5.20 & 2.40 & 46.56 & 47.60 & 34.79 & 52.61 & 38.34 & \underline{45.30} & 34.24 & 52.62 & 37.12 & 42.52 \\
MA-RAG~\citep{wu2026marag} & 4.40 & 4.40 & 45.06 & 46.23 & 34.53 & \underline{55.50} & 38.16 & 38.70 & 30.15 & \underline{55.97} & 37.28 & 37.34 \\
DermAgent~\citep{liu2026dermagent} & \underline{6.70} & 7.10 & 44.77 & 51.62 & 27.71 & 46.52 & 33.82 & 37.66 & 23.68 & 49.25 & 31.54 & 35.72 \\
LCAgent\rx~\citep{hao2026lungcure} & 5.10 & 2.30 & 42.54 & 36.89 & 13.70 & 21.20 & \underline{39.84} & 35.88 & 13.39 & 19.11 & 33.34 & 31.10 \\
\addlinespace[1pt]
\rowcolor{oursrow} \method{}\rx{} (ours) & \textbf{18.20} & \textbf{23.10} & \textbf{64.16} & \textbf{66.73} & \textbf{54.18} & \textbf{63.34} & \textbf{50.28} & \textbf{63.34} & \textbf{50.30} & \textbf{60.69} & \textbf{41.70} & \textbf{54.68} \\
\bottomrule
\end{tabular}}
\caption{\textbf{Results on LungCURE.} Staging, treatment planning with the reference stage, and the end-to-end task from the record, in Chinese (ZH) and English (EN). F1 is the LungCURE plan agreement score and RQ the staging reasoning score. Best in bold, second underlined; shaded rows are ours. $^{\star}$: rules execute the guideline.}
\label{tab:main}
\end{table*}

\subsection{Division of Responsibilities}
\label{sec:boundary}

The core principle of MedDEx is not merely to incorporate a rule executor, but to establish a clear boundary between LLMs and deterministic rules. Formally, clinical record interpretation and guideline execution are constrained as follows:
\begin{equation}
\begin{aligned}
(\hat{s},e)&\in
\left(\prod_{i=1}^{m}
\mathcal{V}_i\right)
\times\mathcal{E},\\
R_{\mathcal{B}}(s)&\subseteq\mathcal{O},
\end{aligned}
\end{equation}
where $\mathcal{E}$ denotes the clinical evidence space. The LLM extracts the facts and evidence required by guideline conditions, while the rule executor handles guideline-defined decisions and derived-condition computations.

Placing the boundary here is what separates MedDEx from an agent that merely calls rules somewhere in its pipeline. A rule executor can only be as reliable as the state it is given: if the model hands it free text and a predicted stage, the model's intermediate errors reach the decision anyway. MedDEx therefore connects the two through structured condition values and assigns every guideline-defined derived computation to the executor. Section~\ref{sec:analysis} measures what each of these choices contributes.

\section{Experiments}
\label{sec:results}

\subsection{Experimental Datasets}
\label{sec:setup}

We evaluate \method{} on three clinical guideline benchmarks (Table~\ref{tab:datasets}, Appendix~\ref{app:external}). \textbf{MedGUIDE}~\citep{li2025medguide} contains multiple-choice questions based on guideline decision trees, evaluated using option accuracy. \textbf{MEGA-CDP}~\citep{chen2026megacdp} requires decision paths and conclusions, evaluated using the official conclusion agreement and path similarity metrics. \textbf{LungCURE}~\citep{hao2026lungcure} contains 1,000 bilingual lung cancer cases in Chinese and English, covering TNM staging (Task~1), treatment planning with the reference stage (Task~2), and end-to-end treatment planning (Task~3). We report exact TNM accuracy, staging reasoning quality (RQ), drug precision, and treatment plan agreement (F1). All methods receive the same guideline knowledge for each case.

\subsection{Compared Methods}

We divide the compared methods into two categories. \textbf{LLM-executed methods} include Direct Prompting, Guideline-RAG, MedAgents~\citep{tang2023medagents}, MDAgents~\citep{kim2024mdagents}, MA-RAG~\citep{wu2026marag}, and DermAgent~\citep{liu2026dermagent}, which employ direct prompting, guideline retrieval, multi-agent collaboration, multi-round retrieval, and planning with self-reflection. \textbf{Rule-executed methods} include \method{} and LCAgent~\citep{hao2026lungcure} on LungCURE. All baselines retain their core designs, with implementation adaptations detailed in Appendix~\ref{app:baselines}.

\subsection{Model Implementation}

On LungCURE, we evaluate Qwen2.5-7B-Instruct~\citep{qwen2025qwen25}, Qwen3-8B~\citep{qwen2025qwen3}, and Llama-3.1-8B-Instruct~\citep{grattafiori2024llama3}. The other two benchmarks use Qwen3-8B and Llama-3.1-8B-Instruct. All models are deployed locally using vLLM~\citep{kwon2023vllm}, with greedy decoding and fixed random seeds, without fine-tuning. Within each comparison, all model components use the same base model. LLM-based evaluations are repeated five times, with method identities hidden from the judge. Routing is additionally scored against public benchmark data only: a keyword parser over the released clinician plans and the released reference TNM. Additional implementation and evaluation details are provided in Appendix~\ref{app:metrics}.

\subsection{Overall Performance}

\paragraph{LLM Execution vs.\ Rule Execution.} Table~\ref{tab:external} compares the two approaches under the same guideline knowledge and base models. \method{} leads every column, with particularly notable improvements in conclusion agreement on MEGA-CDP. On MedGUIDE the margin is under half a point, because multiple candidate options may remain after guideline execution; what separates the methods there is cost rather than accuracy, as \method{} answers with 2.0 model calls per case against 20.7 for MedAgents (Table~\ref{tab:guideline_cost}). No LLM-executed method consistently performs best across benchmarks.

\paragraph{Results on LungCURE.} Table~\ref{tab:main} shows that \method{} achieves the highest exact TNM accuracy across all six model-language settings and the highest treatment plan agreement in five of them. In contrast, retrieval-augmented, self-reflective, and multi-agent methods do not consistently outperform direct prompting.

\section{Analysis of the LLM--Rule Boundary}
\label{sec:analysis}

\begin{figure*}[t]
\centering
\includegraphics[width=\textwidth]{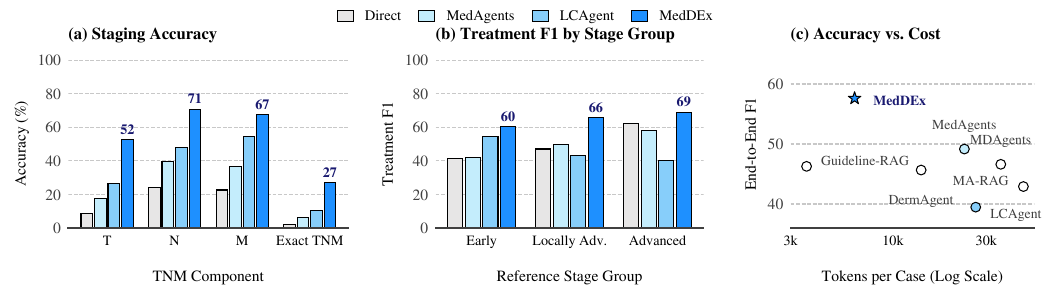}
\caption{\textbf{Downstream effects of the execution boundary.} Qwen2.5-7B, English. (a) Agreement with reference T, N, and M categories and exact TNM. (b) LungCURE plan agreement by reference stage group. (c) End-to-end plan agreement against measured model tokens per case. Scores measure benchmark agreement; the cost protocol is in Appendix~\ref{app:metrics}.}
\label{fig:analysis}
\end{figure*}

This section investigates how the division of responsibilities between LLMs and deterministic rules
affects guideline-based treatment planning. Through component ablations, input interface comparisons,
and efficiency analyses, we examine the contributions of structured patient states, deterministic
guideline execution, and their interaction.

\subsection{Ablation of \method{}}
\label{sec:ablation}

We conduct ablation studies to evaluate the contributions of key components in \method{}. As shown in
Table~\ref{tab:ablation}, each variant modifies one component: replacing rule-based scenario selection
with LLM routing, removing the stage calculator, or providing all nine guidelines instead of the
selected one. Replacing rule-based selection reduces treatment plan agreement by 4.7 points and
end-to-end agreement by 3.2 points. Removing the stage calculator decreases exact TNM accuracy by 18.8
points, while end-to-end agreement drops by only 1.3 points. Since Task~2 uses the reference stage,
its results remain unaffected. Providing all nine guidelines leads to the largest reduction in
treatment agreement (15.7 points), highlighting the importance of selecting relevant guideline content
before generation. These results demonstrate the distinct contributions of deterministic staging,
scenario selection, and guideline-conditioned generation.

\begin{table}[t]
\centering\small
\setlength{\tabcolsep}{1.6pt}
\newcommand{\abcell}[3]{\begin{tabular}[c]{@{}r@{}}#1\\[-2.5pt]{\scriptsize\textcolor{#2}{#3}}\end{tabular}}
\newcommand{\absig}[2]{\abcell{#1}{gainup}{#2}}
\newcommand{\abns}[2]{\abcell{#1}{black!40}{#2}}
\newcommand{\absame}[1]{\abcell{#1}{black!40}{0.0}}  
\begin{tabular}{l Wr WrWr WrWr}
\toprule
& TNM & \multicolumn{2}{c}{Treatment} & \multicolumn{2}{c}{End-to-end} \\
\cmidrule(lr){3-4}\cmidrule(lr){5-6}
Variant & Acc$\uparrow$ & P$\uparrow$ & F1$\uparrow$ & P$\uparrow$ & F1$\uparrow$ \\
\midrule
\rowcolor{oursrow} \method{} (full) & 26.90 & 73.90 & 65.50 & 70.58 & 57.62 \\
\midrule
\quad w/o rule selection & \absame{26.90} & \absig{65.17}{$-$8.7$^*$} & \absig{60.81}{$-$4.7$^*$} & \absig{64.15}{$-$6.4$^*$} & \absig{54.39}{$-$3.2$^*$} \\
\midrule
\quad w/o stage calculator & \absig{8.08}{$-$18.8$^*$} & \absame{73.90} & \absame{65.50} & \abns{70.69}{+0.1} & \absig{56.32}{$-$1.3$^*$} \\
\midrule
\quad w all nine guidelines & \absame{26.90} & \absig{60.27}{$-$13.6$^*$} & \absig{49.82}{$-$15.7$^*$} & \absig{61.24}{$-$9.3$^*$} & \absig{44.89}{$-$12.7$^*$} \\
\bottomrule
\end{tabular}
\caption{\textbf{Ablation results of \method{}.} Values below each score indicate changes from the full model. $^{*}$ denotes a 95\% bootstrap confidence interval excluding zero.}
\label{tab:ablation}
\end{table}

\begin{table}[t]
\centering\small
\setlength{\tabcolsep}{3pt}
\begin{tabularx}{\columnwidth}{@{}>{\raggedright\arraybackslash}X WrWr Wr@{}}
\toprule
& \multicolumn{2}{c}{Plan agreement} & Stage \\
\cmidrule(lr){2-3}\cmidrule(lr){4-4}
Routing input & Exact & Group & consistency \\
\midrule
\rowcolor{oursrow} \method{} (full) & \textbf{67.0} & \textbf{78.9} & \textbf{71.6} \\
\;$-$ stage calculator & 65.3 & 77.9 & 68.5 \\
\;$-$ staging call & 54.5 & 65.5 & 54.3 \\
\;$-$ typed state & 24.7 & 59.6 & 46.4 \\
\bottomrule
\end{tabularx}
\caption{\textbf{Effect of routing inputs on rule execution.} Results on Qwen2.5-7B, English (\%). The rule executor remains unchanged while input components are progressively removed. Plan agreement is measured against scenarios extracted from clinician plans. Stage consistency is evaluated against reference TNM categories.}
\label{tab:ladder}
\end{table}

\begin{table}[t]
\centering\small
\setlength{\tabcolsep}{6pt}
\renewcommand{\arraystretch}{1.25}
\begin{tabular}{l WcWc}
\toprule
& \multicolumn{2}{c}{Stage computed by} \\
\cmidrule(lr){2-3}
Patient state & the model & the rules \\
\midrule
free text & 46.2$^{*}$ & 48.9$^{*}$ \\
structured & 55.9$^{*}$ & \textbf{57.6} \\
\bottomrule
\end{tabular}
\caption{\textbf{Effect of patient state representation and staging responsibility.} End-to-end plan agreement (F1) on Qwen2.5-7B, English. The rule router remains unchanged across all settings. $^{*}$ indicates a 95\% bootstrap confidence interval excluding zero relative to the full system.}
\label{tab:factorial}
\end{table}

\begin{table}[!tb]
\centering\small
\setlength{\tabcolsep}{3pt}
\renewcommand{\arraystretch}{1.25}
\begin{tabular}{@{}l WrWr WrWr@{}}
\toprule
& \multicolumn{2}{c}{MedGUIDE} & \multicolumn{2}{c}{MEGA-CDP} \\
\cmidrule(lr){2-3}\cmidrule(lr){4-5}
Method & Calls & Sec & Calls & Sec \\
\midrule
Single LLM & \textbf{1.0} & \textbf{0.9} & \textbf{1.0} & 49.7 \\
Guideline-RAG & \textbf{1.0} & 2.3 & \textbf{1.0} & \textbf{8.0} \\
MedAgents & 20.7 & 152.3 & 20.5 & 211.1 \\
MDAgents & 19.4 & 303.9 & 17.1 & 399.3 \\
MA-RAG & \underline{1.1} & 1.5 & 5.9 & 164.9 \\
DermAgent & 2.3 & 29.7 & 2.9 & 307.6 \\
\midrule
\rowcolor{oursrow} \method{}\rx{} (ours) & 2.0 & \underline{1.1} & \underline{2.3} & \underline{15.0} \\
\bottomrule
\end{tabular}
\caption{\textbf{Computational cost per case across clinical guideline benchmarks.} Results on Qwen3-8B using the same GPU and concurrency settings. We report the average number of model calls and execution time (seconds) per case. Best in bold, second underlined. $^{\star}$ indicates rule-based guideline execution.}
\label{tab:guideline_cost}
\end{table}

\begin{table}[t]
\centering\footnotesize
\setlength{\tabcolsep}{2.5pt}
\renewcommand{\arraystretch}{1.1}
\begin{tabularx}{\columnwidth}{@{}>{\raggedright\arraybackslash}X WrWr Wr Wr@{}}
\toprule
& \multicolumn{2}{c}{All} & Non-met. & Metastatic \\
\cmidrule(lr){2-3}\cmidrule(lr){4-4}\cmidrule(lr){5-5}
Scenario chosen by & Exact & Grp. & Exact & Exact \\
\midrule
scenario list only (\S\ref{sec:ablation}, Q1) & 38.3 & 54.4 & 69.4 & 4.9 \\
\;+ guideline entry rules, verbatim & 59.5 & 75.2 & 69.2 & 49.1 \\
\;+ chain of thought & 58.3 & 94.2 & 64.2 & 52.1 \\
\;+ 9 worked examples & 35.4 & 93.3 & 58.2 & 11.0 \\
\;+ self-consistency, 5 samples & 36.1 & 93.3 & 58.2 & 12.5 \\
\;+ constrained decoding (no CoT) & 12.0 & 51.7 & 23.3 & 0.0 \\
\midrule
\rowcolor{oursrow} deterministic executor, same input & 79.9 & 94.2 & 73.5 & 86.8 \\
\bottomrule
\end{tabularx}
\caption{\textbf{Improving LLM-based scenario routing does not close the gap.} All methods receive identical patient states and reference stages. We report overall scenario agreement and exact agreement across non-metastatic and metastatic cases (\%).}
\label{tab:router}
\end{table}

\subsection{What the Executor Depends On}
\label{sec:matched}

Although deterministic rules improve guideline execution, their effectiveness depends on the quality of
the inputs they receive. To investigate this dependency, we progressively remove the components
responsible for producing routing inputs while keeping the rule executor unchanged. As shown in
Table~\ref{tab:ladder}, removing the stage calculator, separate staging call, and structured patient
state reduces exact scenario agreement from 67.0\% to 65.3\%, 54.5\%, and 24.7\%, respectively. Stage
consistency similarly decreases from 71.6\% to 46.4\%. In the final setting, the executor relies on a
free-text patient profile and LLM-predicted disease stages. These results demonstrate that
deterministic execution alone is insufficient to ensure reliable guideline decisions. Instead, the
accuracy and structure of the inputs provided to the rule executor play a critical role in downstream
decision performance.

\subsection{Structure and Execution Are One Design}

To further distinguish the contributions of structured patient states and rule-based staging, we
conduct a factorial ablation study while keeping the rule router and treatment generation module
unchanged.

As shown in Table~\ref{tab:factorial}, we compare free-text and structured patient states under both
LLM-based and rule-based staging. Replacing free-text profiles with structured states improves
end-to-end plan agreement by an average of 9.0 F1 points, while assigning staging computation to
deterministic rules provides an additional average improvement of 2.1 points. The full \method{}
configuration achieves the highest F1 score of 57.6.

These results indicate that the structured state interface contributes more substantially to
downstream performance than the staging arrangement alone. They further highlight the importance of
combining structured patient representations with a clear division of responsibilities between LLMs
and deterministic rules.

\subsection{Can Prompts Replace Rules?}
\label{sec:router}

We further investigate whether stronger prompting strategies can achieve performance comparable to
deterministic rule execution. As shown in Table~\ref{tab:router}, we progressively enhance LLM-based
scenario selection by incorporating guideline entry rules, chain-of-thought reasoning, worked examples,
self-consistency, and constrained decoding, while keeping patient states and reference stages
unchanged. The worked examples are labeled using the rule executor's decisions.

Adding guideline entry rules improves exact scenario agreement from 38.3\% to 59.5\%, but additional
reasoning strategies fail to consistently improve performance. Even the best LLM-based configuration
remains 20.4 points below deterministic rule execution (79.9\%). The gap is particularly pronounced in
metastatic cases, where the highest LLM agreement reaches 52.1\%, compared with 86.8\% for
deterministic rules.

These results indicate that stronger prompting and additional reasoning cannot reliably replace
deterministic execution of guideline-defined decision logic in the evaluated settings.

\subsection{Staging, Treatment, and Cost}

Finally, we investigate how the division of responsibilities between LLMs and rule executors affects
downstream decision performance and computational efficiency. As shown in Figure~\ref{fig:analysis},
\method{} improves agreement with reference TNM categories and treatment plans across disease stage
groups while requiring approximately four to seven times fewer tokens than multi-agent approaches.

Table~\ref{tab:guideline_cost} further demonstrates the efficiency advantages of \method{}. On
MedGUIDE, \method{} achieves performance comparable to the strongest baselines with only 2.0 model
calls and 1.1 seconds per case, compared with 20.7 calls and 152.3 seconds for MedAgents. On MEGA-CDP,
\method{} achieves higher conclusion agreement with only 2.3 model calls and 15.0 seconds per case,
compared with 20.5 calls and 211.1 seconds for MedAgents.

Although \method{} is not the fastest method in every setting, it achieves a favorable balance between
decision accuracy and computational cost. These results further demonstrate that executing
guideline-defined decision logic through deterministic rules is more efficient than relying on
repeated LLM reasoning.

\FloatBarrier
\section{Conclusion}

In this paper, we propose \method{}, a guideline-based treatment planning framework that combines
structured patient states with deterministic rule execution. Experiments demonstrate that \method{}
improves guideline decision agreement while reducing inference costs. Further analyses highlight the
importance of clearly dividing responsibilities between LLMs and rules, suggesting that additional
reasoning does not always lead to better guideline execution. More generally, when a decision procedure is
already written down and its decision factors can be named and checked, executing it is both more
reliable and cheaper than asking a model to re-derive it.

\FloatBarrier
\clearpage
\section*{Limitations}
\label{sec:limitations}

Our evaluation is limited to three relatively small language models and a fixed set of clinical
guidelines. Although \method{} improves guideline decision agreement, exact TNM accuracy remains
limited, and benchmark agreement does not establish clinical reliability. Furthermore, the reference
scenarios are derived from clinician plans rather than independent clinical adjudication. Further
validation is needed to assess clinical applicability and generalizability.

\section*{Ethical considerations}

This study uses publicly available clinical benchmarks without attempting to re-identify patients or
disclose sensitive information. AI coding assistants were used to draft executable guideline rules,
which were subsequently reviewed and tested by the authors. \method{} is intended for research
purposes only and requires further clinical validation before real-world deployment.

\bibliography{refs}
\clearpage
\appendix
\section{Benchmarks and Evaluation}
\label{app:external}

\begin{table}[htbp]
\centering\small
\setlength{\tabcolsep}{3pt}
\begin{tabularx}{\columnwidth}{@{}l W >{\raggedright\arraybackslash}X W r W >{\raggedright\arraybackslash}X@{}}
\toprule
Dataset & Guideline & Size & Task \\
\midrule
MedGUIDE & 53 author trees & 7,747 & multiple-choice decision \\
MEGA-CDP & 440 guideline trees & 7,458 & decision path and conclusion \\
LungCURE & 9 NCCN/CSCO scenarios & 1{,}000 & staging, plan, end to end \\
\bottomrule
\end{tabularx}
\caption{\textbf{Guideline benchmarks.} LungCURE cases are available in Chinese and English.}
\label{tab:datasets}
\end{table}

\subsection{Benchmark datasets and evaluation}

\paragraph{MedGUIDE~\citep{li2025medguide}.}
We use all 7,747 public multiple-choice questions, covering 53 distinct clinical decision trees. Each
item gives a short patient description and a set of candidate recommendations, and the task is to
select the option the decision tree leads to. Because the correct answer is among the given options,
this benchmark measures whether a method reaches the right final decision rather than how it reasoned
there. It is scored by option accuracy.

\paragraph{MEGA-CDP~\citep{chen2026megacdp}.}
We use all 7,458 official test cases, together with their 440 clinical guidelines and decision trees.
The dataset requires both a decision path and a final conclusion. Each case gives a patient record and
the guideline it falls under, and the model has to derive the decision path step by step from that
guideline before stating the recommendation. Scoring follows the official rubric: conclusion agreement
for the recommendation, and an optimal-transport similarity between the predicted path and the
reference path. Because the decision path is itself scored, this benchmark assesses guideline
execution more directly.

\paragraph{LungCURE~\citep{hao2026lungcure}.}
The dataset contains real lung cancer records together with the plan the treating clinician actually
wrote, in Chinese and English for the same patient. It has three tasks: assigning the TNM stage from
the record, recommending treatment when the reference stage is supplied, and doing both end to end. It
is scored with the official plan-agreement judge. Of the three benchmarks, LungCURE is the only one
whose reference is a clinician's own decision rather than a guideline endpoint.

\subsection{Metrics and reporting}
\label{app:metrics}

MedGUIDE uses exact option accuracy.

For MEGA-CDP, conclusion agreement uses the released rubric and parser, is averaged over repeated judge calls, and is mapped to 0--100. Path similarity uses the released pairwise scores and optimal transport.

Cases where the model fails stay in the denominator. We report only metrics that cover all cases and pass the scoring checks.

\paragraph{LungCURE scores.}

We follow the official LungCURE scoring: drug precision, the plan agreement score the benchmark calls \emph{F1}, and the staging reasoning score, each with the released prompts and rubric. We report them as percentages under the benchmark's own names, and estimate paired differences with 5{,}000 case-level bootstrap resamples. Table~\ref{tab:proxy} additionally reports a drug overlap score computed from normalized drug names without an LLM judge.

\paragraph{Routing reference.}

Section~\ref{sec:setup} defines the three routing metrics; this section states how the parser behind
the first one works. It is a fixed program with no language model. It reads the diagnosis line of the
released clinician plan when the plan has one, together with treatment keywords.

\FloatBarrier
\section{Baseline Implementations}
\label{app:baselines}

Each comparison uses the same base model for every role in a run. No method sees test-labeled examples. Within a dataset, all methods receive the same source knowledge; a rebuilt rule pack is never given to \method{} alone. We serve all methods with vLLM~\citep{kwon2023vllm} and use greedy decoding.

\paragraph{Direct prompting.} The model receives the task prompt only.

\paragraph{Guideline-RAG.} The system retrieves guideline passages related to the patient and gives them to the model when it writes the treatment plan.

\paragraph{MedAgents.} Several clinical specialists assess the case, exchange views, and refine a shared recommendation, following the released sequence of specialist analysis, synthesis, and verification with five question-domain specialists and two option-domain specialists.

\paragraph{MDAgents.} A moderator estimates how complex the case is and then chooses between a single-clinician answer and specialist collaboration.

\paragraph{MA-RAG.} The system compares candidate decisions, retrieves more evidence based on their disagreements, and then refines the recommendation, with the released intrinsic-entropy ranking over eight candidates and eight rounds, its conflict queries, and majority aggregation.

\paragraph{DermAgent.} A planner chooses the steps and tools, an executor carries out the analysis, and a critic checks the result and suggests fixes. Its tools read the patient record and the guideline rather than images.

\paragraph{LCAgent.} The system extracts evidence from the record, builds a patient profile, performs staging and rule-based scenario selection, and writes a plan from the selected guideline.

\FloatBarrier
\section{\method{} Implementation}
\label{app:rules}

\subsection{Rule sources}

The rule execution code was written by two coding assistants, GPT-6-Astra and Claude Opus 5.5. The
LungCURE scenario rules are built from the relevant NCCN and CSCO guidelines. The MedGUIDE and
MEGA-CDP rules are compiled directly from the released clinical guideline decision trees, without
using test answers.

We check the rule
implementations against the original guidelines through structural checks, comparison with the source
material, and author review; these checks do not constitute independent clinical validation.

For MedGUIDE and MEGA-CDP, \method{} extracts the values of the guideline conditions. The executor runs the compiled conditions, following the design of each benchmark. The model then writes the answer from the retained outcomes and the shared guideline. The model still judges whether a clinical condition holds, and it may resolve an ambiguous outcome from the record and the guideline. We credit the rule sources to the benchmark authors.

\begin{algorithm}[htbp]
\caption{\method{} decision process}
\label{alg:meddex}
\small\raggedright
\begin{algorithmic}[1]
\Require record $x$; rule executor $R_G$ developed from guideline $G$ with conditions $P$; LLM $\mathcal{M}$
\Ensure answer $y$
\State $s \gets \mathcal{M}(x, P)$ \Comment{LLM: condition values and raw evidence}
\State $o \gets R_G(s)$ \Comment{rules: derived conditions, then the decision graph}
\State $y \gets \mathcal{M}(x, s, G_o)$ \Comment{LLM: answer within decision $o$}
\State \Return $y$
\end{algorithmic}
\end{algorithm}

\subsection{Patient information}

The patient state records the facts needed for staging and treatment choice: histology, performance status, molecular alterations, PD-L1, surgery, prior treatment, progression, contraindications, and the extent of metastatic disease.

The extraction prompt allows unknown values for histology, driver status, performance status, and PD-L1, so the writer can be told that a condition was not determined rather than being given a default.

\subsection{Computing the stage}

The calculator follows the AJCC 8th-edition staging rules. It starts from tumor-size thresholds; if the record reports qualifying invasion or additional lung nodules, it raises the T category. Invasion of the chest wall, parietal pleura, or pericardium can raise T to T3. Invasion of the mediastinum or major adjacent structures, or a nodule in a different lobe, can raise T to T4.

For N, the calculator takes the highest confirmed lymph node group: hilar or intrapulmonary, mediastinal or subcarinal, then supraclavicular, scalene, or contralateral nodes. For M, it uses confirmed distant disease after filtering out uncertain and benign descriptions. It excludes regional nodes and separates spread within the chest, a single distant lesion, and multiple distant lesions.

\subsection{Selecting a guideline scenario}

Table~\ref{tab:scenarios} lists the rules in execution order. For metastatic disease, the router first checks for a driver alteration and then for progression after prior treatment. For the remaining cases, it checks whether the disease is oligometastatic. 

For non-metastatic disease, the router first checks whether curative surgery is complete, then uses the node and tumor categories to separate unresectable locally advanced, potentially resectable, and resectable disease.

\begin{table}[htbp]
\centering\small
\setlength{\tabcolsep}{3pt}
\begin{tabularx}{\columnwidth}{@{}l X@{}}
\toprule
Scenario & Entry condition (checked in order) \\
\midrule
Adv.\ driver$+$, later line & M1, driver$+$, prior therapy and progression \\
Adv.\ driver$+$, first line & M1, driver$+$ \\
Oligometastatic & M1, oligometastatic, $\le$5 lesions, first line \\
Adv.\ driver$-$, later line & M1, prior therapy and progression \\
Adv.\ driver$-$, first line & M1 \\
Postoperative adjuvant & M0, curative resection done \\
Unresectable locally adv. & M0, N3 \\
Potentially resectable & M0, N2 or T3/T4 \\
Resectable & M0, otherwise \\
\bottomrule
\end{tabularx}
\caption{\textbf{Guideline scenarios and the rules that select them.}}
\label{tab:scenarios}
\end{table}

\subsection{Converting a guideline into executable rules}

A rule pack is a transcription of material that already exists, not a new clinical guideline. For
MedGUIDE and MEGA-CDP the source is the decision tree released with the benchmark; for LungCURE it is
the scenario definitions in the published NCCN and CSCO guidelines. In both cases the conversion
turns the document into the three parts Section~\ref{sec:process} defines: the conditions $P$, the
derived-condition rules $\mathcal{F}$, and the decision graph $G$.

The conversion preserves the structure of the source rather than reorganising it. Section headings
stay headings. Conditions keep their original AND/OR relations. Notes on timing and eligibility stay
attached to the treatments they qualify. The result is checked against the source document, and it adds
no clinical content of its own.

\FloatBarrier
\section{Further Analyses}

\begin{table}[htbp]
\centering\small
\setlength{\tabcolsep}{2pt}
\begin{tabular*}{\columnwidth}{@{\extracolsep{\fill}}l rrrr rrrr@{}}
\toprule
& \multicolumn{4}{c}{Treatment (T2)} & \multicolumn{4}{c}{End-to-end (T3)} \\
\cmidrule(lr){2-5}\cmidrule(lr){6-9}
Method & Calls & In & Out & s & Calls & In & Out & s \\
\midrule
Guideline-RAG & 1.0 & 2.9 & 0.59 & 0.8 & 1.0 & 2.9 & 0.70 & 0.8 \\
DermAgent & 4.6 & 12.0 & 1.80 & 2.3 & 4.6 & 12.1 & 1.82 & 2.1 \\
LCAgent\rx{} & 1.8 & 3.5 & 0.47 & 2.4 & 7.8 & 22.3 & 4.18 & 4.5 \\
MedAgents & 9.1 & 19.2 & 3.95 & 4.9 & 9.0 & 19.0 & 4.23 & 5.2 \\
MDAgents & 10.7 & 27.8 & 7.35 & 8.8 & 10.7 & 28.0 & 7.63 & 9.1 \\
MA-RAG & 11.9 & 31.6 & 5.17 & 6.3 & 14.3 & 39.6 & 6.85 & 8.3 \\
\midrule
\rowcolor{oursrow} \method{}\rx{} & 2.0 & 3.6 & 0.31 & 0.4 & 3.0 & 5.7 & 0.68 & 0.9 \\
\bottomrule
\end{tabular*}
\caption{\textbf{Cost per case.} All methods use the same model, hardware, and concurrency. Calls, input and output tokens (thousands), and average time cover the full pipeline. $^{\star}$: rules execute the guideline.}
\label{tab:efficiency}
\end{table}

\begin{table}[htbp]
\centering\footnotesize
\setlength{\tabcolsep}{2pt}
\begin{tabularx}{\columnwidth}{@{}>{\raggedright\arraybackslash}X W>{\raggedright\arraybackslash}X WrWrWrWr Wr@{}}
\toprule
& & \multicolumn{4}{c}{Accuracy (\%)} & \\
\cmidrule(lr){3-6}
\textbf{Stage} & \textbf{Expl.} & T$\uparrow$ & N$\uparrow$ & M$\uparrow$ & TNM$\uparrow$ & RQ$\uparrow$ \\
\midrule
\rowcolor{oursrow} Calculator & Rule (ours) & \textbf{52.49} & \textbf{70.79} & \textbf{67.43} & \textbf{26.90} & \textbf{70.78} \\
Calculator & Model's reply & \textbf{52.49} & \textbf{70.79} & \textbf{67.43} & \textbf{26.90} & \underline{46.49} \\
Model & Model's reply & \underline{27.05} & \underline{36.47} & \underline{44.15} & \underline{8.08} & 44.60 \\
\bottomrule
\end{tabularx}
\caption{\textbf{Staging and its explanation.} The stage comes from the stage calculator or the model's own answer; the explanation comes from the rule that set each category or the model's own answer. RQ is the judge's reasoning score for T, N, and M. Best in bold, second underlined; the shaded row is the full system.}
\label{tab:rq}
\end{table}

\begin{table*}[htbp]
\centering\small
\begin{tabular}{l WrWrWrWr WrWrWrWr}
\toprule
& \multicolumn{4}{c}{Treatment (Task 2)} & \multicolumn{4}{c}{End-to-end (Task 3)} \\
\cmidrule(lr){2-5}\cmidrule(lr){6-9}
\textbf{Method} & P$\uparrow$ & R$\uparrow$ & F1$\uparrow$ & \#drugs & P$\uparrow$ & R$\uparrow$ & F1$\uparrow$ & \#drugs \\
\midrule
Direct prompting & 48.8 & 38.1 & 39.7 & 3.8 & 47.3 & 43.0 & 42.3 & 4.3 \\
Guideline-RAG & 42.2 & 40.1 & 38.2 & 3.9 & 42.2 & 41.8 & 39.0 & 4.1 \\
MedAgents & 49.1 & 36.4 & 38.3 & 3.6 & 41.1 & \underline{44.7} & 38.9 & 5.6 \\
MDAgents & 46.9 & 35.6 & 36.8 & 3.5 & 44.1 & 38.8 & 37.6 & 4.1 \\
MA-RAG & \underline{52.2} & \underline{39.4} & \underline{41.2} & 3.5 & \underline{52.2} & 38.8 & \underline{41.1} & 3.5 \\
DermAgent & 45.6 & 39.0 & 38.9 & 3.9 & 44.8 & 39.8 & 39.2 & 4.2 \\
LCAgent\rx{} & 31.8 & 32.8 & 30.0 & 3.3 & 37.4 & 39.3 & 35.2 & 4.3 \\
\addlinespace[1pt]
\rowcolor{oursrow} \method{}\rx{} (ours) & \textbf{60.4} & \textbf{46.6} & \textbf{49.0} & 2.8 & \textbf{65.2} & \textbf{50.3} & \textbf{52.8} & 3.5 \\
\bottomrule
\end{tabular}
\caption{\textbf{Drug metrics without an LLM.} We compare the drug names in each plan with the clinician's plan and report precision, recall, F1, and drugs per plan; no language model is used for scoring. Best in bold, second underlined; shaded rows are ours. $^{\star}$: rules execute the guideline.}
\label{tab:proxy}
\end{table*}

\begin{table*}[htbp]
\centering\small
\renewcommand{\arraystretch}{1.12}
\definecolor{casered}{HTML}{E0301E}
\newcommand{\cyes}{\textcolor{gainup}{\ding{51}}}
\newcommand{\cno}{\textcolor{casered}{\ding{55}}}
\newcommand{\cpart}{\textcolor{gaindn}{$\sim$}}
\newcommand{\cfact}[1]{\textbf{\textcolor{gainup}{#1}}}
\begin{tabularx}{\textwidth}{@{}>{\raggedright\arraybackslash}p{3.1cm}W>{\raggedright\arraybackslash}XW>{\raggedright\arraybackslash}p{5.3cm}Wc@{}}
\toprule
\multicolumn{4}{@{}p{\textwidth}@{}}{\textit{Record:} 1.8\,cm lung adenocarcinoma, \cfact{already resected} by lobectomy; \cfact{1 of 2 hilar nodes positive}; \cfact{EGFR L858R} mutation.}\\
\multicolumn{4}{@{}p{\textwidth}@{}}{\textit{Clinician:} stage T1bN1M0 $\rightarrow$ postoperative adjuvant therapy $\rightarrow$ osimertinib for $\ge$3 years.}\\
\midrule
\textbf{Method} & \textbf{How the patient was read} & \textbf{Plan} & \\
\midrule
MedAgents & No setting chosen & Chemotherapy or ``any EGFR-TKI'' & \cno \\
LCAgent & Positive node missed (T1bN0) & Osimertinib & \cpart \\
\rowcolor{oursrow} \textbf{\method{}} & Resected: yes; driver: EGFR L858R; stage T1bN1M0 (exact) $\Rightarrow$ \emph{postoperative adjuvant} & Adjuvant osimertinib for $\ge$3 years & \cyes \\
\bottomrule
\end{tabularx}
\caption{\textbf{Case study.} Each row shows how a system read the patient and the plan it wrote; key facts in the record are in blue. \cyes: matches the clinician; \cpart: partly matches, with a staging error; \cno: differs from the reference setting. \method{} records the key facts, computes the reference stage, and selects postoperative adjuvant treatment.}
\label{tab:case}
\end{table*}

\subsection{Cost protocol}

Table~\ref{tab:efficiency} measures each method on a shared set of evaluation cases, with the same model server, GPU type, and request concurrency. We start from the patient record and count the successful model calls and all input and output tokens needed to produce the result.

We include the full \method{} pipeline. If a variant reuses a cached patient state from another run, we also charge it for the call that produced that state. LCAgent runs its own frozen implementation in a separate job under the same protocol. Every method uses Qwen2.5-7B on one NVIDIA A100 80GB PCIe with 16 concurrent requests. The \method{} Task~3 rows include all three model calls, including staging; the other rows count the patient-state call reused from the Task~3 $k{=}3$ run.

The average token count of this case set is close to that of the full evaluation set, while the differences between methods are much larger, so the results support comparing calls and tokens. The reported time is total batch time divided by the number of cases, at fixed concurrency. It is not the waiting time of a single request. Under this protocol, Guideline-RAG is still the fastest method per case, so we do not claim a latency advantage for \method{}.

\subsection{Staging decision and its explanation}
\label{app:components}

Table~\ref{tab:rq} separates the stage prediction from its explanation. It compares the model's own reply with a stage computed from extracted evidence plus an explanation of which rule set each category. This lets us evaluate the decision and the explanation separately.

\subsection{Drug overlap without an LLM judge}

Table~\ref{tab:proxy} checks whether the gains also appear in drug overlap metrics that do not rely on an LLM judge. These tables describe the evaluated pipeline with its missing-value defaults; they do not evaluate a system with stricter input checks or abstention.

\subsection{Case studies}
\label{app:cases}

Because each stage of the pipeline is recorded separately, a plan can be read back as the sequence of
decisions that produced it. Table~\ref{tab:case} follows one case through that sequence. The record
states a resected 1.8\,cm adenocarcinoma with one positive hilar node and an EGFR L858R mutation.
\method{} registers all three facts, the calculator returns T1bN1M0, the executor reaches the
postoperative adjuvant branch, and the writer stays inside it. The baselines diverge at an identifiable
point rather than everywhere: one never commits to a treatment setting, and another reads the positive
node as absent and routes the case as node-negative. The same reading applies when the final plan is
wrong, which is the practical value of the decomposition: the disagreement can be attributed to the
extracted state, to the computed stage, to the selected scenario, or to the written text, instead of to
the system as a whole.

\end{document}